\documentclass[lettersize,conference]{IEEEtran}
\IEEEoverridecommandlockouts

\usepackage{cite}
\usepackage{amsmath,amssymb,amsfonts}
\usepackage{algorithmic}
\usepackage{graphicx}
\usepackage{textcomp}
\usepackage{xcolor}
\usepackage{url}
\usepackage{booktabs}
\usepackage{multirow}
\usepackage{pifont}
\usepackage[caption=false,font=footnotesize]{subfig}
\newcommand{\cmarkbold}{\ding{52}}
\newcommand{\xmarkbold}{\ding{56}}
\def\BibTeX{{\rm B\kern-.05em{\sc i\kern-.025em b}\kern-.08em
    T\kern-.1667em\lower.7ex\hbox{E}\kern-.125emX}}

\begin{document}

\title{OmniVLoc: Topo-metric Visual Localization for Omnidirectional Cameras}

\author{Zhuo Yu\textsuperscript{1}, Shuyang Zhang\textsuperscript{1,$\dagger$},
  Huihua Zhu\textsuperscript{1}, Wenjia Ouyang\textsuperscript{1,2},
  Zuo Zuo\textsuperscript{3},
  Jianhao Jiao\textsuperscript{4}, \\
  Yilong Zhu\textsuperscript{5},
  Xiaopin Zhong\textsuperscript{1}, and Zongze Wu\textsuperscript{1}
  \thanks{\textsuperscript{1} Zhuo Yu, Shuyang Zhang, Huihua Zhu, Wenjia Ouyang,
    Xiaopin Zhong, and Zongze Wu are with the College of Mechatronics
    and Control Engineering, Shenzhen University, Shenzhen, Guangdong, 518060,
    China.~(Corresponding author: Shuyang Zhang)}
  \thanks{\textsuperscript{2} Wenjia Ouyang is with the College of Computer Science
    and Software Engineering, Shenzhen University, Shenzhen, Guangdong, 518060, China.}
  \thanks{\textsuperscript{3} Zuo Zuo is with College of Artificial Intelligence,
    Xi'an Jiaotong University, Xi'an, Shaanxi, 710049, China.}
  \thanks{\textsuperscript{4} Jianhao Jiao is with the Robot Perception and Learning Lab, Intelligent Robotics,
    Department of Computer Science, University College London, Gower Street, WC1E 6BT, London, UK.}
  \thanks{\textsuperscript{5} Yilong Zhu is with the Department of Electronic and Computer Engineering,
    Hong Kong University of Science and Technology, Hong Kong.}
}

\maketitle

\begin{abstract}
Omnidirectional cameras provide surround-view information, significantly overcoming 
the field-of-view constraints for traditional visual localization systems. 
However, existing localization systems fail to fully exploit the characteristics 
of omnidirectional cameras, struggling to deliver accurate and robust localization 
performance amidst scene content variations and illumination-induced appearance changes.
In this work, we propose OmniVLoc, a surround-view localization system based on 
a topo-metric framework.
By integrating state-of-the-art global place recognition and monocular depth estimation methods, 
we introduce a multi-constrained visual localization paradigm.
By solving a scale-aware pose graph and performing precise spherical bundle adjustment, 
we accurately recover the metric-level pose for each query frame.
Furthermore, a hierarchical verification mechanism is designed to filter out erroneous 
outputs across different stages, effectively bolstering the overall system robustness. 
Extensive experiments on both public and our custom datasets validate the superior 
accuracy and reliability of OmniVLoc. 
To benefit the community, our source code and dataset will be made publicly available upon publication.

\end{abstract}

\section{Introduction}
\begin{figure}
  \centering
  \includegraphics[width=0.95\columnwidth]{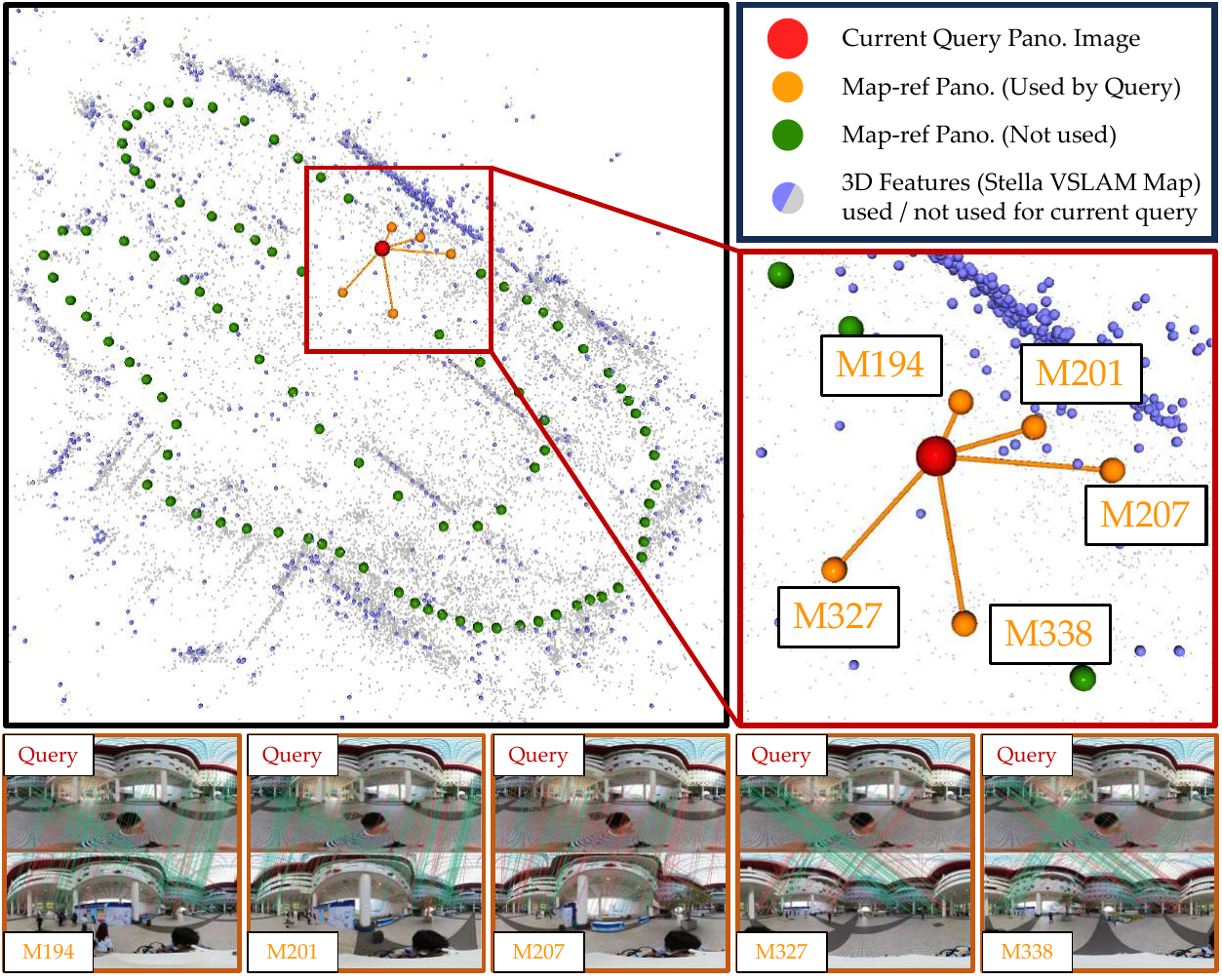}
  \caption{Comparison between our topo-metric map and traditional 3D feature map.
  Traditional 3D mapping relies on multi-view triangulation and suffers from geometric
  degeneracy and complex updates.
  Our topological map treats pose-annotated reference panoramas (orange) as independent
  database keys for the query (red), eliminating rigid geometric constraints, enabling effortless
  map management and continuous updates.}
  \label{fig_cover}
\end{figure}

Accurate metric visual localization is a fundamental prerequisite 
for autonomous navigation in mobile robot systems~\cite{yin2025tro,yu2026tase}. 
While traditional structure-based systems achieve high precision by establishing 
2D-3D correspondences against pre-built 3D maps, they face significant scalability 
bottlenecks.
Constructing and storing dense 3D points, local descriptors, and co-visibility graphs 
incur prohibitive overhead in large-scale environments~\cite{camposeco2019cvpr}. 
Furthermore, maintaining these 3D maps over extended periods is exceedingly difficult; 
feature matching and triangulation are highly sensitive to viewpoint shifts and 
illumination-induced appearance variations. 
Consequently, dynamic environmental changes and map obsolescence drastically 
degrade localization performance~\cite{vidal2018ral,zhang2026sj}, rendering 
traditional 3D map-based paradigms inadequate for large-scale deployments and 
lifelong autonomy~\cite{toft2020tpami} (as conceptually compared in Fig.~\ref{fig_cover}).

In contrast, recent topo-metric localization paradigms~\cite{wang2024iclr,jiao2025icra,jiao2026arxiv}
replace dense point clouds with sparse topological image nodes,
significantly reducing storage and maintenance costs while performing local metric
recovery at these nodes.
However, existing topo-metric methods face two major limitations.
First, they typically rely on conventional cameras with a narrow field of view~(FOV). 
This necessitates redundant multi-view storage to capture the full scene during mapping, 
and the limited visual context severely exacerbates perceptual aliasing during localization~\cite{yin2025tro2}.
Second, local metric recovery in these systems is fragile in complex environments: 
stereo-vision-based methods are bounded by physical baselines, while traditional 
monocular approaches suffer from severe scale ambiguity and drift, compromising 
absolute pose accuracy.

To address these challenges, we propose OmniVLoc, a surround-view localization system 
based on a topo-metric framework.
By introducing this ultra-wide FOV camera, we effectively overcome the issues of
perceptual aliasing and map redundancy.
To achieve precise metric recovery without relying on pre-built 3D geometry or 
stereo baselines, we introduce a multi-constrained visual localization paradigm.
Specifically, we leverage a visual foundation model~(VFM) for monocular depth estimation 
and formulate the query-to-map registration as geometric hypotheses regarding the 
unknown depth scale.
By solving a scale-aware pose graph and performing precise spherical bundle adjustment, 
OmniVLoc accurately recovers the metric-level pose for each query frame.
Additionally, a hierarchical verification mechanism is designed to filter out erroneous outputs, 
effectively bolstering system robustness against illumination and scene variations.
Our main contributions are summarized as follows:
\begin{itemize}
  \item A continuous metric localization framework for surround-view topo-metric maps.
        This framework obviates the need for pre-built 3D map points or depth maps,
        providing a unified formulation for precise metric localization using 
        omnidirectional cameras.
  \item A scale-aware factor graph optimization scheme. 
        It leverages VFM-derived monocular depth with scale uncertainty to construct 
        multi-hypothesis 3D-2D visual constraints, tightly coupled with temporal 
        motion priors to recover accurate absolute scales and poses.
  \item A hierarchical verification mechanism based on multiple observation
        and motion priors, ensuring system stability under severe perceptual aliasing 
        and feature degradation.
  \item Extensive experiments on both public and our custom datasets, which validate 
        the superior accuracy and robustness of OmniVLoc compared to state-of-the-art baselines. 
\end{itemize}

\section{Related Works}
Visual localization is a fundamental task in robotics, aiming to estimate 
the camera pose within a known map via visual observations.
Existing methods can be categorized into \textit{structure-based},
\textit{learning-based}, and \textit{topological} visual localization.
\subsection{Structure-based Visual Localization}
Structure-based visual localization estimates absolute camera poses by establishing
2D-3D correspondences against a pre-built 3D map.
Traditional systems like ORB-SLAM~\cite{campos2021tro} rely on Bag-of-Words for
coarse localization and hand-crafted features for pose optimization.
To improve robustness in complex environments, recent pipelines like Hierarchical 
Localization~(HLoc)~\cite{sarlin2019cvpr} and PixLoc~\cite{sarlin2021cvpr} integrate 
learned descriptors and feature-metric optimization.
Some work~\cite{he2024icra} even fuse offline LiDAR point clouds as geometric priors 
to push the absolute accuracy limits.
Nevertheless, these methods heavily rely on global 3D structural maps, incurring
prohibitive storage costs and rendering them brittle in dynamic environments~\cite{saputra2018csur},
where scene changes necessitate time-consuming re-triangulation and global optimization.

\subsection{Learning-based Visual Localization}
Learning-based approaches encode scene representations directly into neural network parameters. 
Absolute pose regression~(APR) networks like PoseNet~\cite{kendall2015iccv} directly 
regress 6 degrees of freedom~(DoF) poses, while scene coordinate regression~(SCR) 
methods like ACE Zero~\cite{brachmann2024eccv} map images to 3D coordinates without 
ground-truth poses. 
Recent advancements like Reloc3r~\cite{dong2025cvpr} improve zero-shot generalization 
via relative pose regression. 
While Novel View Synthesis techniques~\cite{liu2025iclr, gao2025tase} are frequently 
adopted for post-optimization pose refinement, their computational overhead hinders 
real-time large-scale deployment. 
Furthermore, compressing vast scene details into limited network weights inevitably 
leads to catastrophic forgetting~\cite{wang2021iccv}. 
Updating the map to reflect physical environmental changes necessitates computationally 
expensive network retraining~\cite{maggio2023icra}, severely limiting their scalability 
for lifelong robot autonomy.

\subsection{Topological and Topo-metric Visual Localization}
Topological visual localization represents environments as lightweight graphs, 
avoiding the massive geometric maps required by structure-based methods.
While visual place recognition~(VPR)~\cite{izquierdo2024cvpr,berton2025cvpr}
enables efficient scene redirection, it yields only coarse localization.
To achieve precise 6-DoF estimation, recent works have shifted towards topo-metric paradigms.
Methods like LiteVLoc~\cite{jiao2025icra}, OpenNavMap~\cite{jiao2026arxiv}, and 
FreeReg~\cite{wang2024iclr} achieve metric registration directly between query 
and reference images without global 3D maps.

However, existing topo-metric methods predominantly rely on perspective cameras. 
Their limited field of view~(FOV) requires dense orientation sampling during mapping 
and suffers from severe perceptual aliasing during localization. 
Moreover, their metric recovery heavily relies on stereo baselines or monocular depth accuracy, 
which are baseline-bounded and scale-ambiguous in complex environments.

In this work, we introduce omnidirectional cameras into the topo-metric framework 
to formulate OmniVLoc.
The holistic surround-view effectively overcomes FOV limitations and mitigates 
perceptual aliasing~\cite{guo2026360ral}. 
By integrating a VFM for omnidirectional depth recovery and formulating a scale-aware 
factor graph with multi-frame constraints, our system successfully recovers precise, 
absolute-scale metric poses without relying on pre-built 3D geometry or stereo setups.

\section{Methodology}
\subsection{Problem Formulation and System Overview}
\label{sect_problem_overview}

\begin{figure*}
  \centering
  \includegraphics[width=0.95\textwidth]{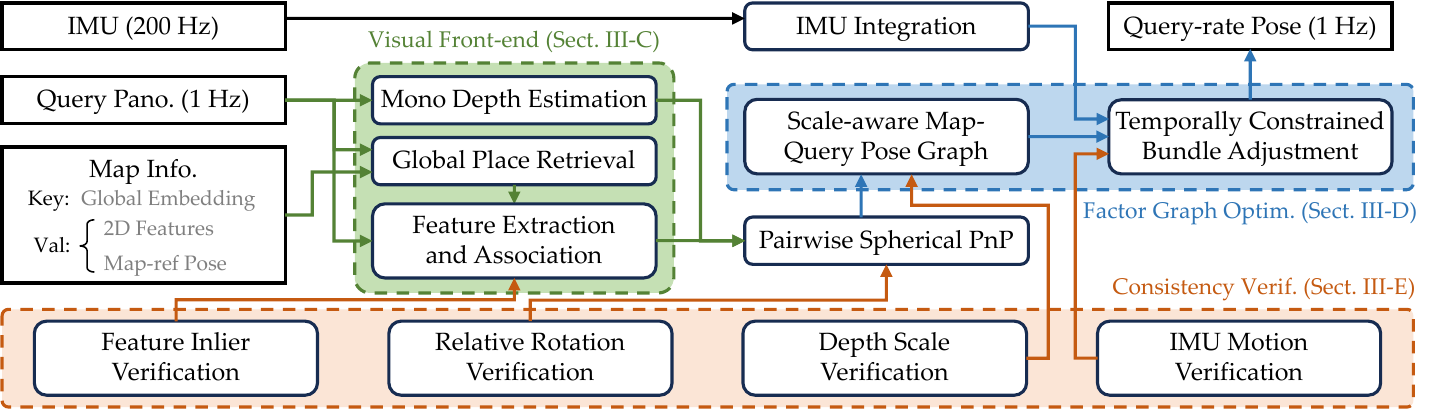}
  \caption{Overview of our system. System front-end processes omnidirectional images 
  to obtain monocular depth with scale uncertainty, global place candidates, 
  and their feature associations with the query frame. 
  The back-end utilizes a factor graph to eliminate scale uncertainty, 
  followed by pose optimization with temporal constraints and bundle adjustment. 
  Across the entire pipeline, we enforce strict consistency checks on intermediate outputs 
  to avoid system degradation. }
  \label{fig_system_overview}
\end{figure*}

\subsubsection{Problem Formulation}
We define a pose-annotated image-node map as
$\mathbf{\mathcal{M}}=\left\{\left(\mathbf{I}_{i}^m,\mathbf{T}_{i}^{m},\mathbf{g}_i\right)\right\}_{i=1}^{N}$,
where $\mathbf{I}_{i}^m$, $\mathbf{T}_{i}^{m}$, and $\mathbf{g}_i$ are the $i$-th reference
panorama, its camera-to-world pose, and global descriptor, respectively.
Note that raw images are not stored in practice; we only retain their 2D feature 
coordinates and descriptors.
During mapping, we extract a panoramic sequence from a metric mapping system (e.g., LIVO).
To ensure map efficiency, we sparsify the sequence by discarding redundant frames 
based on spatial distance ($<\tau_d$) and descriptor similarity ($>\tau_s$).
Unlike traditional structure-based maps, $\mathcal{M}$ is purely topo-metric, 
containing neither 3D landmarks nor co-visibility graphs.

In the localization phase, given an incoming sequence of query images 
$\mathbf{\mathcal{I}}_{0:t}=\{\mathbf{I}_k^q\}_{k=0}^{t}$,
our goal is to estimate the trajectory
$\mathbf{\mathcal{X}}_{0:t}=\{\mathbf{T}_{k}^{q}\}_{k=0}^{t}$
conditioned on inter-frame motion observations $\mathcal{O}_{1:t}$ (e.g., Inertial 
Measurement Unit (IMU) measurements).
We formulate this as a Maximum A Posteriori~(MAP) estimation
\begin{equation}
  \mathcal{X}_{0:t}^{*}=\arg
  \underset{\mathcal{X}_{0:t}}{\max}\;
  p\!\left(\mathcal{X}_{0:t}\mid
  \mathcal{I}_{0:t},\mathcal{O}_{1:t},\mathcal{M}\right).
  \label{eq_problem_formulation}
\end{equation}
Evaluating Eq.~\eqref{eq_problem_formulation} over the entire map is
computationally prohibitive. 
Therefore, we decouple the problem into a VPR-based coarse retrieval stage
(yielding a candidate subset $\mathcal{J}_k$ of size $N_{\mathrm{ret}}$) 
and a metric factor-graph optimization stage:
\begin{equation}
  \begin{aligned}
    \{\mathcal{X}_{0:t}^{*},\mathcal{J}_{0:t}^{*}\}
    =\arg\underset{\mathcal{X}_{0:t},\mathcal{J}_{0:t}}{\max}\quad
     & \underbrace{p\!\left(\mathcal{X}_{0:t}\mid
         \mathcal{I}_{0:t},\mathcal{O}_{1:t},
         \mathcal{M}_{\mathcal{J}_{0:t}}\right)}_{\text{Metric factor-graph estimation}}              \\
    \times 
    & \underbrace{\prod_{k=0}^{t}\prod_{j\in\mathcal{J}_k}
               p\!\left(j\mid \mathbf{I}_k^q,\mathcal{M}\right)}_{\text{VPR-based coarse retrieval}}.
  \end{aligned}
  \label{eq_hierarchical_formulation}
\end{equation}

\subsubsection{System Overview}
As illustrated in Fig.~\ref{fig_system_overview}, OmniVLoc operates in two stages. 
The \textit{front-end} retrieves the top-$N_{\mathrm{ret}}$ map nodes, extracts monocular depth 
with scale uncertainty, and establishes 2D-3D correspondences. 
The \textit{back-end} optimizes a scale-aware factor graph to recover continuous, 
absolute-scale 6-DoF poses without relying on pre-built 3D geometry.
\textit{Consistency checks} span both modules, eliminating incorrect intermediate outputs, 
thereby guaranteeing the robustness of the system.

\subsection{Preliminaries: Omnidirectional Geometry}
\label{sect_omni_geo}
Omnidirectional images are stored as equirectangular projection~(ERP) image, whose
image-plane distortion varies with latitude. 
More details on spherical camera model can be found in \cite{huang2022360vo}.
We map all 2D pixels $\mathbf{u}=[u,v]^\top$ to unit-sphere bearings 
$\mathbf{f}(\mathbf{u})\in\mathbb{S}^{2}$ and formulate all geometric errors in 
the spherical domain. 

For a correspondence between a query 3D point $\mathbf{P}_{k,j}^{q}\in\mathbb{R}^{3}$
and its matched map bearing  $\mathbf{f}_{i,j}^{m}$, let 
$\mathbf{T}_{ik}^{mq}=(\mathbf{T}_{i}^{m})^{-1}\mathbf{T}_{k}^{q}
=[\mathbf{R}_{ik}^{mq},\mathbf{t}_{ik}^{mq}]$ be the relative transformation 
from the query frame $k$ to the map frame $i$.
We define the predicted bearing $\widehat{\mathbf{f}}_{i,j}^{m}$ and the spherical 
vector residual $\mathbf{e}_{i,k,j}$ as:
\begin{equation}
  \widehat{\mathbf{f}}_{i,j}^{m}
     =\frac{\mathbf{R}_{ik}^{mq}\mathbf{P}_{k,j}^{q}
      +\mathbf{t}_{ik}^{mq}}
    {\left\|\mathbf{R}_{ik}^{mq}\mathbf{P}_{k,j}^{q}
      +\mathbf{t}_{ik}^{mq}\right\|_2}, \quad
    \mathbf{e}_{i,k,j} = \mathbf{f}_{i,j}^{m}\times\widehat{\mathbf{f}}_{i,j}^{m}.
  \label{eq_spherical_pnp_residual}
\end{equation}
The residual $\mathbf{e}_{i,k,j}$ effectively approximates the angular error.
The robust spherical Perspective-n-Point (PnP) estimate is then given by minimizing
$\sum\rho\!\left(\|\mathbf{e}_{i,k,j}\|_2^2\right)$,
where $\rho(\cdot)$ is a robust loss function.

\subsection{Panoramic Visual Front-end}
\label{sect_visual_frontend}
The front-end retrieves plausible map nodes, establishes query--map feature
correspondences, and combines them with query depth to form the 3D-to-bearing
constraints used by spherical PnP.

\subsubsection{Global Topological Retrieval}
\label{subsect_topo_retrieval}
Given a query panorama $\mathbf{I}_k^q$, we employ
MegaLoc~\cite{berton2025cvpr} to extract its global descriptor $\mathbf{g}_k^q$ 
and rank all map nodes by descriptor cosine similarity.
Since a panorama may overlap multiple nearby nodes, we retain the top
$N_{\mathrm{ret}}$ candidates rather than only the highest-ranked
image~\cite{jiao2025icra}; all candidates are passed to local matching
($N_{\mathrm{ret}}=5$ in our setup).

\subsubsection{CubeMap-based Feature Association}
\label{subsect_feature_association}
We adopt SuperPoint~\cite{detone2018cvpr} and LightGlue~\cite{lindenberger2023iccv} 
for feature extraction and association.
Since these models target perspective images, we decompose each panorama into six 
CubeMap faces to avoid ERP distortion.
SuperPoint features extracted per face are back-projected and concatenated in
the ERP domain, after which LightGlue matches the query against each candidate.

\subsubsection{Omnidirectional Depth Estimation}
\label{subsect_depth_estimation}
Since $\mathcal{M}$ lacks pre-triangulated landmarks for PnP, we use Depth Any 
Panoramas~(DAP)~\cite{lin2026cvpr} to predict a dense depth map
$\widetilde{\mathbf{D}}_k$ for the query. 
Although DAP claimed to provide metric depth, we observe a bounded scale drift 
in its real-world predictions.
This drift linearly propagates to the local 3D points and the PnP translation. 
Thus, we introduce a positive per-query scale factor $s_k$:
\begin{equation}
  \mathbf{P}_{k,j}^{q}=s_k\widetilde{\mathbf{P}}_{k,j}^{q},
  \quad
  \mathbf{t}_{ik}^{mq}=s_k\widehat{\mathbf{t}}_{ik}^{mq}.
  \label{eq_depth_scale_model}
\end{equation}
All query--map registrations at time $k$ share scale $s_k$, 
which is subsequently optimized in the factor graph.

\begin{figure}
  \centering
  \includegraphics[width=0.9\columnwidth]{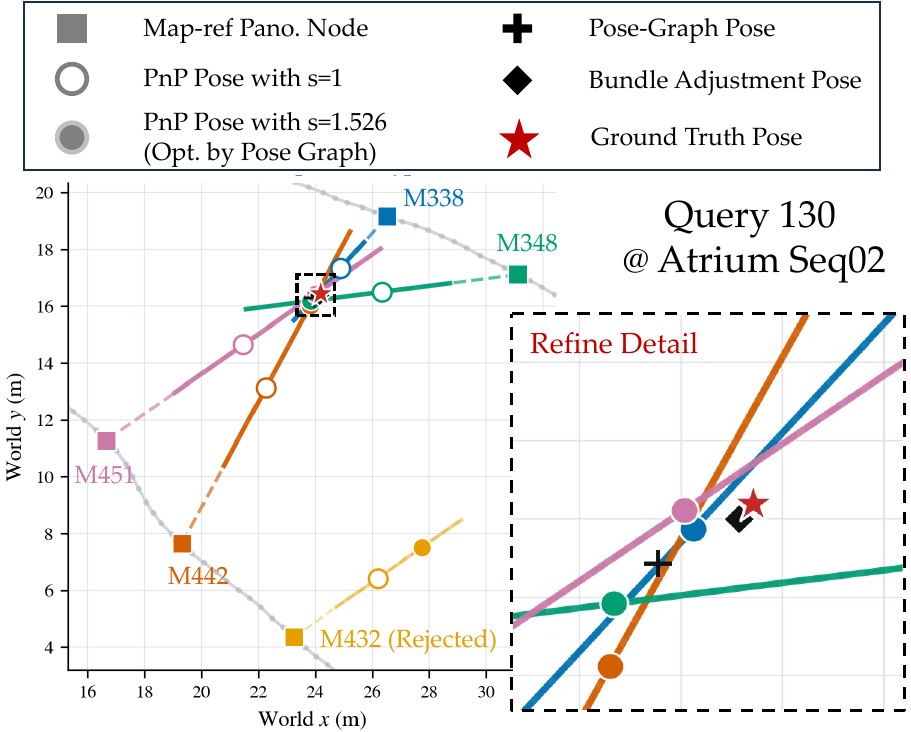}
  \caption{Demonstration of our pose optimization process.
  The dashed lines illustrate the possible query positions yielded by a single PnP 
  operation under unknown scales, while the solid lines indicate the valid candidates 
  restricted by scale boundaries of $[0.5, 2.0]$. 
  }
  \label{fig_scale_estimation}
\end{figure}

\subsection{Scale-aware Factor-Graph Estimation}
\label{sect_factor_graph}
The metric factor-graph term in Eq.~\eqref{eq_hierarchical_formulation}
would grow with all historical states and measurements. 
For a bounded online update, we adopt a first-order Markov assumption. 
Estimation at time $k$ becomes
\begin{equation}
  \mathcal{X}_k^{*}
  =\underset{\mathcal{X}_k}{\arg\max}\;
  p\!\left(
  \mathcal{X}_k\mid
  \mathcal{I}_k,\mathcal{O}_k,
  \mathcal{M}_{\mathcal{J}_k},\mathcal{X}_{k-1}
  \right),
  \label{eq_markov_update}
\end{equation}
where the optimization state is augmented as
$\mathcal{X}_k=\{\mathbf{T}_k^q,s_k\}$ to account for the shared DAP scale.
We realize Eq.~\eqref{eq_markov_update} with two stages~(see Fig.~\ref{fig_scale_estimation}).
A pure-visual pose graph first estimates the current pose and the shared scale 
from pairwise PnP observations. 
Its output initializes spherical bundle adjustment~(BA), which combines point-level 
bearing factors with $\mathcal{O}_k$ and $\mathcal{X}_{k-1}$. 

\subsubsection{Scale-Aware Visual Pose Graph}
\label{subsect_pose_graph}
Each verified map node $i\in\mathcal{J}_k^{\mathrm{ver}}$ provides a relative PnP
observation
$\widehat{\mathbf{T}}_{ik}^{mq}=[\widehat{\mathbf{R}}_{ik}^{mq},
    \widehat{\mathbf{t}}_{ik}^{mq}]$, while the current state predicts
$\mathbf{T}_{ik}^{mq,\mathrm{pred}}=(\mathbf{T}_i^m)^{-1}\mathbf{T}_k^q$.
Parameterizing the shared scale as $s_k=\exp(\alpha_k)$, we define the normalized
rotation, translation, and scale-prior residuals:
\begin{equation}
  \begin{aligned}
    \mathbf{r}_{ik}^{R}
     & =
    \frac{
      \left[\operatorname{Log}\!\left(
      (\widehat{\mathbf{R}}_{ik}^{mq})^{\top}
      \mathbf{R}_{ik}^{mq,\mathrm{pred}}\right)\right]^{\vee}}
    {\sigma_R}, \\
    \mathbf{r}_{ik}^{t}
     & =
    \frac{
      \mathbf{t}_{ik}^{mq,\mathrm{pred}}
      -\exp(\alpha_k)\widehat{\mathbf{t}}_{ik}^{mq}}
    {\sigma_t}, \\
    r_k^s & =\frac{\alpha_k}{\sigma_s}.
  \end{aligned}
  \label{eq_pose_graph_residuals}
\end{equation}
Here, $(\cdot)^{\vee}$ maps an element of $\mathrm{SO}(3)$ or
$\mathrm{SE}(3)$ to its Lie vector space, and $\sigma_R$, $\sigma_t$,
and $\sigma_s$ normalize the residuals. 
The visual pose graph optimizes $ \{\mathbf{T}_k^{q,\mathrm{pg}},\alpha_k^*\}$
by minimizing the Huber loss over these residuals.
Excluding temporal and inertial factors keeps this scale estimate determined 
solely by the current multi-map visual geometry.

\subsubsection{Temporally Constrained Spherical Bundle Adjustment}
\label{subsect_spherical_ba}
The pose graph output initializes a local BA.
Each correspondence initializes a world-frame landmark as
$\mathbf{P}_l^{w,0}=\mathbf{R}_k^{q,\mathrm{pg}}
  (s_k^*\widetilde{\mathbf{P}}_{k,l}^q)+\mathbf{t}_k^{q,\mathrm{pg}}$.
For any observation frame $a\in\{q_k,m_i\}$, the spherical bearing residual is
$\mathbf{r}_{a,l}^{\mathrm{bear}}=\mathbf{f}_{a,l}\times
  \operatorname{normalize}[\mathbf{R}_a^\top
    (\mathbf{P}_l^w-\mathbf{t}_a)]$. 
Map poses remain fixed while the current query pose and local landmarks are optimized.

Furthermore, consecutive queries and integrated IMU measurements provide a relative 
motion prior $\Delta\widehat{\mathbf{T}}_{k-1,k}^{o}$.
With the previous optimized pose $\overline{\mathbf{T}}_{k-1}^{q}$ fixed, 
the temporal residual is
\begin{equation}
  \mathbf{r}_{k-1,k}^{\mathrm{temp}}
  =
  \left[\operatorname{Log}
  \left(
  (\Delta\widehat{\mathbf{T}}_{k-1,k}^{o})^{-1}
  (\overline{\mathbf{T}}_{k-1}^{q})^{-1}
  \mathbf{T}_k^q
  \right)\right]^{\vee}.
  \label{eq_ba_temporal_residual}
\end{equation}
Defining the BA state as $\boldsymbol{\Theta}_k=\{\mathbf{T}_k^q,\{\mathbf{P}_l^w\}\}$, 
the final objective minimizes the joint cost of bearing observations and temporal priors:
\begin{equation}
  \boldsymbol{\Theta}_k^*
  =
  \underset{\boldsymbol{\Theta}_k}{\arg\min}\sum_{(a,l)\in\mathcal{V}_k}
  \rho\!\left(
  \|\mathbf{r}_{a,l}^{\mathrm{bear}}\|_{\boldsymbol{\Sigma}_b}^2
  \right) + \|\mathbf{r}_{k-1,k}^{\mathrm{temp}}\|_{\boldsymbol{\Sigma}_o}^2,
  \label{eq_spherical_ba}
\end{equation}
where $\mathcal{V}_k$ contains the query and map bearing observations. 
Fixing $s_k^*$ during BA prevents scale--structure gauge ambiguity. 
A looser temporal covariance $\boldsymbol{\Sigma}_o$ regularizes motion without 
overriding the multi-map visual evidence.

\begin{figure}
  \centering
  \includegraphics[width=0.95\columnwidth]{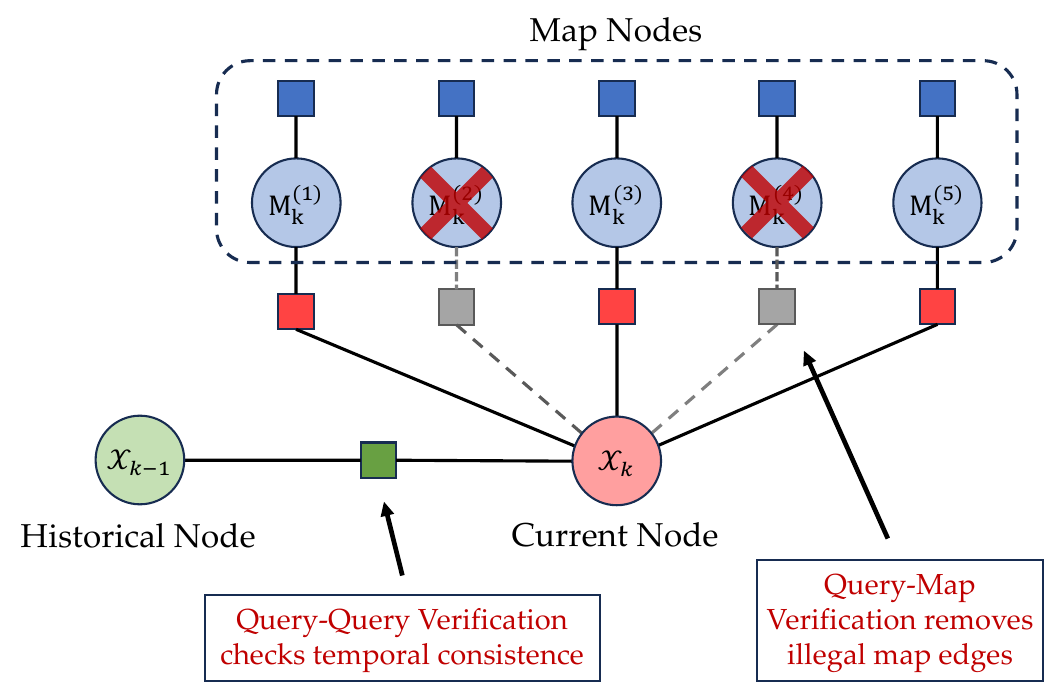}
  \caption{Demonstration of our temporal factor graph and our consistency verification.}
  \label{fig_factor_graph}
\end{figure}

\subsection{Hierarchical Consistency Verification}
\label{sect_consistency_verification}
Cross-time appearance changes and repeated structures can produce incorrect
retrievals and geometrically false matches that corrupt the factor graph. 
We implement a two-level verification (as illustrated in Fig.~\ref{fig_factor_graph}) 
to ensure only healthy constraints enter the factor graph ($\mathcal{J}_{k}^{\mathrm{ver}} \subset \mathcal{J}_{k}$).

\subsubsection{Query--Query Verification}
\label{subsect_query_query_verification}
We validate short-term visual motion between $\mathbf{I}_{k-1}^q$ and $\mathbf{I}_k^q$
before it is used as the temporal factor in Eq.~\eqref{eq_spherical_ba}.
We reject image pairs with insufficient matches or RANSAC inliers, 
a low inlier ratio, or a large median spherical bearing residual.

We also estimate motion in both query directions. 
Consistent forward and inverse-backward PnP solutions form a strong edge; 
a single valid direction is retained with enlarged covariance, while inconsistent 
solutions are rejected.
If the discrepancy between the integrated gyroscope rotation
$\Delta\mathbf{R}_{k-1,k}^{g}$ and the visual PnP rotation 
$\Delta\widehat{\mathbf{R}}_{k-1,k}^{qq}$ exceeds a threshold $\tau_g$, 
the visual edge is deemed unreliable, and the system falls back to purely 
inertial motion without using visual evidence.

\begin{figure}
  \centering
  \includegraphics[width=0.95\columnwidth]{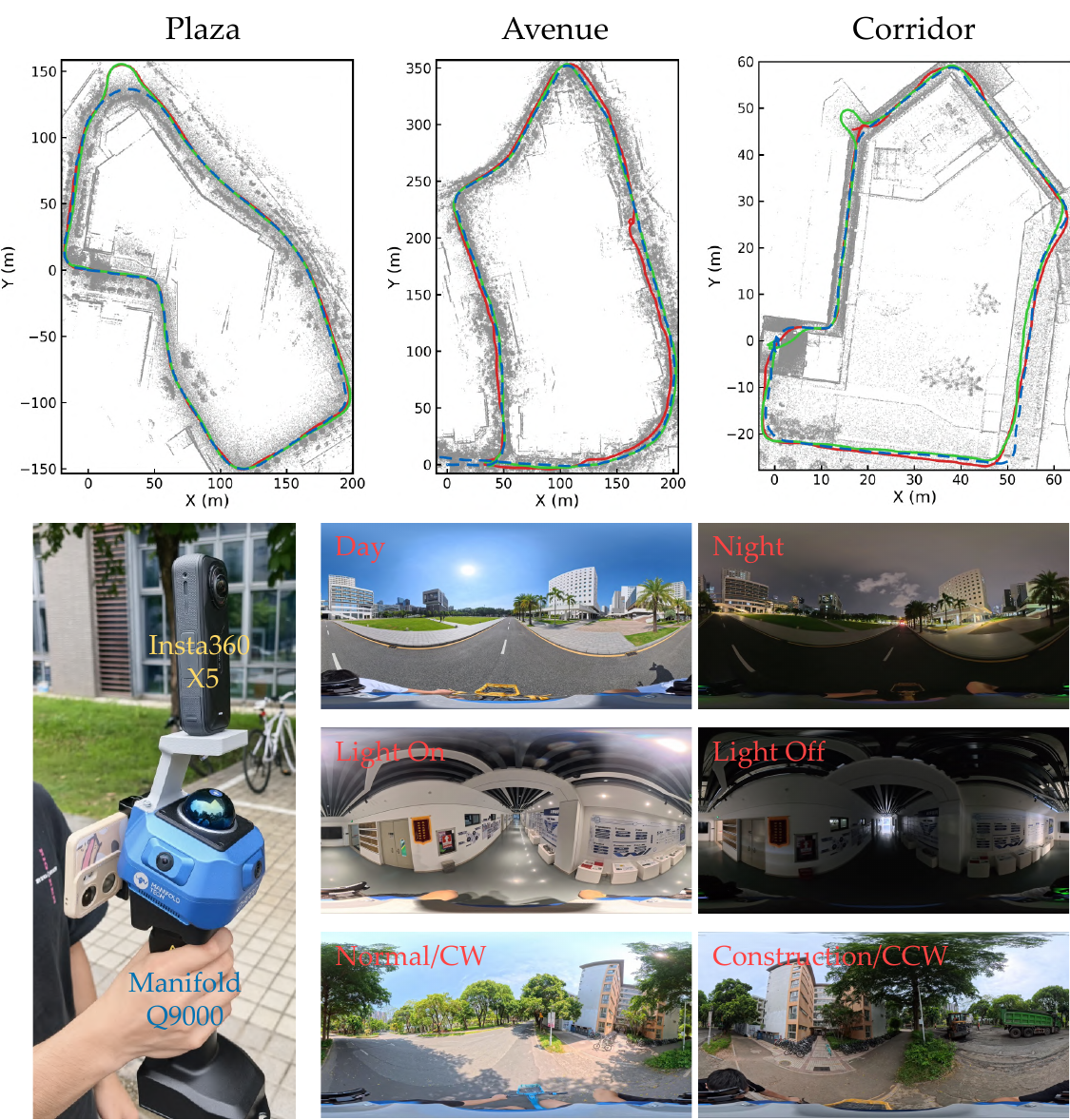}
  \caption{Our custom dataset and capturing devices.
  \textit{Plaza} exhibits extreme day-to-night appearance shifts in an outdoor setting.
  \textit{Corridor} involves a narrow indoor space with varying illumination.
  \textit{Avenue} introduces repetitive foliage alongside ongoing construction events.}
  \label{fig_dataset}
\end{figure}

\subsubsection{Query--Map Verification}
\label{subsect_query_map_verification}
Each retrieved query--map pair first undergoes the same state-independent
quality checks on matches, spherical PnP support, and bearing residual.
Afterwards, a retrieved map node is rejected if its orientation deviates significantly 
from the gyro-propagated prior.
This gate is disabled during cold start or unreliable query--query motion to 
preserve relocalization.

More importantly, since all PnP translations share the uncertain depth scale
$s\in\mathcal{S}=[s_{\min},s_{\max}]$, each candidate defines a valid position ray 
(or segment) in the world frame:
\begin{equation}
  \boldsymbol{\ell}_{ik}(s)
    = \mathbf{t}_i^m+s\mathbf{R}_i^m
    \widehat{\mathbf{t}}_{ik}^{mq}
    = \mathbf{p}_{ik}^{0}+s\mathbf{d}_{ik}.
\end{equation}
Valid map candidates must reach a spatial consensus. 
We find the optimal scale $s_k^{c}$ and intersection center $\mathbf{c}_k$ by 
maximizing the inlier support $w_{ik}$ across intersection rays:
\begin{equation}
  \begin{aligned}
    \{s_k^{c},\mathbf{c}_k,\mathcal{J}_k^{\mathrm{ver}}\}
     & = \arg\underset{s,\mathbf{c},\mathcal{J}}{\max}
    \sum_{i\in\mathcal{J}}w_{ik},                          \\
     & \mathrm{s.t.}\quad \|\boldsymbol{\ell}_{ik}(s)-\mathbf{c}\|_2<\tau_p.
  \end{aligned}
  \label{eq_scale_line_consensus}
\end{equation}
Pairwise segment intersections generate scale hypotheses, which are scored by 
support and compactness.
Historical segments are transported by validated query motion and contribute
only down-weighted, capped support; every valid cluster must include a current
query--map constraint.
The consensus $s_k^c$ initializes $\alpha_k$ in the visual pose graph, 
whose refined scale initializes the BA in Sect.~\ref{subsect_spherical_ba}.

\begin{table}[t]
  \caption{Supported Features of the Baseline Methods}
  \label{tab_supp_feats}
  \centering
  \resizebox{\columnwidth}{!}{%
    \begin{tabular}{@{}lcccl@{}}
      \toprule
      Method & ERP & IMU        & Loc. Mode  & Map Representation            \\ \midrule
      Stella~\cite{sumikura2019acmmm}
             & \cmarkbold & \xmarkbold & \cmarkbold & 2D\&3D Features + BoW         \\
      HLoc~\cite{sarlin2019cvpr}
             & \xmarkbold & \xmarkbold & \cmarkbold & 2D\&3D Features + NetVLAD     \\
      ACE0~\cite{brachmann2024eccv}
             & \xmarkbold & \xmarkbold & \cmarkbold & Network Param.                \\
      Ours
             & \cmarkbold & \cmarkbold & \cmarkbold & 2D Features + MegaLoc         \\ \bottomrule
    \end{tabular}%
  }
\end{table}

\section{Experiment}
\subsection{Datasets}
\label{sect_datasets}
\subsubsection{360Loc Dataset}
We primarily evaluate OmniVLoc on the 360Loc dataset~\cite{huang2024cvpr}, 
which comprises four campus environments with $14$ sequences, covering campus 
scenes with repetitive structures and perceptual aliasing. 
Each scene provides a daytime mapping sequence together with several daytime and 
nighttime localization sequences, recorded at approximately $3$~Hz.
Since 360Loc does not provide IMU measurements, we generate simulated IMU
readings from its ground-truth camera trajectory, with white noise and random walks
for bias, following the EuRoC-ADIS16448 noise model.

\subsubsection{Custom Dataset} 
We also collect a custom campus dataset, supporting a higher query-image rate of $24$~Hz. 
Our acquisition rig (illustrated in Fig.~\ref{fig_dataset}) rigidly combines an 
Insta360 X5 panoramic camera with a Manifold Q9000 commercial mapping device, 
which comprises a Mid-360 LiDAR and three fisheye cameras. 
Its multi-sensor mapping system provides sub-centimeter localization accuracy. 
We use its estimated trajectory as the localization ground truth and calibrate 
the extrinsic transformation between the Insta360 X5 and the Q9000 LiDAR rig, 
allowing the reference poses to be expressed in the panoramic-camera frame.
Similar to the 360Loc protocol, each scene contains a daytime mapping sequence and 
separate daytime and nighttime localization sequences. 
We collect $3$ scenes with $9$ sequences with complementary localization challenges, 
such as open-space geometric sparsity, illumination changes, and severe appearance repetition.

For each scene in both datasets, we construct the panoramic image-node map from daytime 
mapping sequences using our rule defined in Sect.~\ref{sect_problem_overview}. 
We set the distance threshold to $\tau_d=5.0$~m and the descriptor similarity threshold 
to $\tau_s=0.7$. 

\begin{table*}[t]
  \caption{Localization accuracy on the 360Loc and custom datasets. 
  Each entry reports mean/median ATE.
  '-' represents initialization failure or ATE more than $10$.}
  \label{tab_quantitative_performance}
  \centering
  \newlength{\tablecmidrulegap}
  \setlength{\tablecmidrulegap}{\aboverulesep}
  \addtolength{\tablecmidrulegap}{\cmidrulewidth}
  \addtolength{\tablecmidrulegap}{\belowrulesep}
  \begin{tabular}{@{}c|cc|*{5}{c}|cc@{}}
    \toprule
    \multirow{2}{*}[-.5\tablecmidrulegap]{Dataset} & \multirow{2}{*}{Scene} & \multirow{2}{*}{Seq.}
     & \multicolumn{5}{c|}{Single-frame Localization}
     & \multicolumn{2}{c}{Sequence-based Localization} \\
    \cmidrule(lr){4-8}\cmidrule(lr){9-10}
     & & & HLoc-Opt~\cite{sarlin2019cvpr} & HLoc-Fus~\cite{sarlin2019cvpr} & ACE0-Opt~\cite{brachmann2024eccv} & ACE0-Fus~\cite{brachmann2024eccv} & Ours-NoSeq & Stella~\cite{sumikura2019acmmm} & Ours-Full \\ \midrule
    \multirow{14}{*}[-1.5\tablecmidrulegap]{360Loc} & \multirow{4}{*}{Atrium}
     &   D1 & 2.966 / 0.346 & 2.510 / 0.707 & \textbf{1.304} / 0.780 & 1.583 / 1.168 & 2.613 / \textbf{0.321} & 0.320 / 0.282 & \textbf{0.079} / \textbf{0.054}\\
     & & D2 & 2.096 / 0.912 & 1.698 / 0.879 & 1.866 / 0.999 & 2.556 / 1.472 & \textbf{0.344} / \textbf{0.096} & 1.049 / 0.908 & \textbf{0.231} / \textbf{0.111}\\
     & & N1 & 2.743 / 1.017 & 3.120 / 1.147 & - / -       & - / -       & \textbf{2.071} / \textbf{0.193} & - / -       & \textbf{0.407} / \textbf{0.166}\\
     & & N2 & 0.988 / 0.984 & 1.007 / 0.914 & - / -       & - / -       & \textbf{0.322} / \textbf{0.143} & 0.956 / 0.957 & \textbf{0.309} / \textbf{0.155}\\ \cmidrule(lr){2-10}
     & \multirow{3}{*}{Concourse}
     &   D1 & 0.570 / 0.234 & 0.338 / 0.236 & 1.406 / 0.746 & 1.603 / 0.940 & \textbf{0.030} / \textbf{0.019} & 2.116 / 1.442 & \textbf{0.029} / \textbf{0.018}\\
     & & D2 & 5.316 / 1.049 & 5.472 / 1.017 & \textbf{2.010} / 0.765 & 2.142 / 1.089 & 2.286 / \textbf{0.199} & 2.774 / 2.278 & \textbf{0.066} / \textbf{0.022}\\
     & & N1 & 3.021 / 0.943 & 3.309 / 1.000 & - / -       & - / -       & \textbf{0.213} / \textbf{0.087} & 2.567 / 2.283 & \textbf{0.100} / \textbf{0.086}\\ \cmidrule(lr){2-10}
     & \multirow{4}{*}{Hall}
     &   D1 & 2.170 / 0.497 & 2.242 / 0.501 & 2.138 / 0.884 & 2.644 / 1.219 & \textbf{1.192} / \textbf{0.131} & 0.505 / 0.396 & \textbf{0.049} / \textbf{0.021}\\
     & & D2 & 2.439 / 0.448 & \textbf{2.219} / 0.444 & 5.758 / 1.644 & 5.839 / 2.141 & 2.730 / \textbf{0.302} & 0.599 / 0.487 & \textbf{0.059} / \textbf{0.035}\\
     & & N1 & 8.763 / 0.993 & 7.397 / 0.963 & - / -       & - / -       & \textbf{3.094} / \textbf{0.398} & - / -       & \textbf{0.705} / \textbf{0.306}\\
     & & N2 & 3.558 / 1.254 & -           & - / -       & - / -       & \textbf{2.164} / \textbf{0.517} & - / -       & \textbf{0.759} / \textbf{0.443}\\ \cmidrule(lr){2-10}
     & \multirow{3}{*}{Piatrium}
     &   D1 & - / -       & - / -       & 5.203 / 2.269 & 5.476 / 2.563 & \textbf{3.367} / \textbf{0.538} & 5.333 / 4.379 & \textbf{0.215} / \textbf{0.100}\\
     & & D2 & 2.719 / 0.580 & 2.688 / 0.943 & 8.424 / 4.885 & 8.520 / 5.196 & \textbf{1.320} / \textbf{0.262} & 2.637 / 1.149 & \textbf{0.315} / \textbf{0.123}\\
     & & N1 & - / -          & - / -        & - / -        & - / -        & \textbf{6.164} / \textbf{1.487} & - / -      & \textbf{1.842} / \textbf{0.910}\\ \midrule
    \multirow{6}{*}[-\tablecmidrulegap]{Custom} & \multirow{2}{*}{Plaza}
     &   D1 & \textbf{0.351} / 0.236 & 0.381 / 0.277 & - / 9.481 & - / 9.798 & 0.585 / \textbf{0.155} & 1.005 / 0.784 & \textbf{0.438} / \textbf{0.167}\\
     & & N1 & \textbf{0.920} / 0.266 & 1.422 / 0.337 & - / - & - / - & 0.998 / \textbf{0.162} & - / - & \textbf{0.750} / \textbf{0.150}\\ \cmidrule(lr){2-10}
     & \multirow{2}{*}{Corridor}
     &   D1 & 0.344 / \textbf{0.225} & 0.352 / 0.258 & 2.849 / 2.358 & 2.978 / 2.586 & \textbf{0.313} / 0.273 & 0.334 / 0.337 & 0.286 / 0.274\\
     & & N1 & 0.657 / 0.228 & 0.667 / 0.245 & 6.636 / 5.767 & 6.225 / 6.520 & \textbf{0.564} / \textbf{0.140} & - / - & \textbf{0.263} / \textbf{0.140}\\ \cmidrule(lr){2-10}
     & \multirow{2}{*}{Avenue}
     &   D1 & 1.974 / 0.234 & 3.567 / 0.308 & - / - & - / - & \textbf{0.574} / \textbf{0.134} & 1.105 / 1.098 & \textbf{0.214} / \textbf{0.133}\\
     & & N1 & - / 0.222 & - / 0.389 & - / - & - / - & \textbf{0.437} / \textbf{0.190} & - / - & \textbf{0.332} / \textbf{0.189}\\
    \bottomrule
  \end{tabular}%
\end{table*}

\subsection{Baseline Methods and Evaluation Criterion}
Although many open-source visual odometry and SLAM systems are available, few
support global localization with ERP panoramas, and fewer still use inertial
measurements in this setting. 
We select three representative baselines and adapt their inputs to our panoramic setup.
Their supported features are summarized in Table~\ref{tab_supp_feats}:
\begin{itemize}
  \item \textbf{Stella VSLAM~(Stella)}~\cite{sumikura2019acmmm} is a geometric visual
        SLAM system that natively supports ERP images. We use its localization
        mode with a pre-built sparse 3D feature map, representing conventional
        structure-based panoramic localization.
  \item \textbf{Hierarchical Localization~(HLoc)}~\cite{sarlin2019cvpr} is a hierarchical 
        structure-based localization pipeline designed for perspective images.
  \item \textbf{ACE Zero~(ACE0)}~\cite{brachmann2024eccv} is a map-free scene
        coordinate regression method whose scene representation is encoded in
        network parameters.
\end{itemize}
Since HLoc and ACE0 are both designed for perspective images, we apply CubeMap 
adaptations to both methods. 
Each ERP panorama is projected into six perspective faces for mapping and localization. 
The \textbf{Opt} adaptation selects the valid face pose with the strongest geometric verification.
In this mode, ground-truth poses are leaked for optimal pose selection.
The \textbf{Fus} adaptation jointly optimizes the valid pose evidence from all faces 
by a simple pose graph. 

In the localization process, HLoc, ACE0 and our methods are treated 
as online queries and processed at $1$~Hz.
As a visual SLAM system, Stella inherently benefits from high-frequency 
image sequences. 
Therefore, to ensure its optimal performance, we utilized the full-frame-rate image 
sequences across both datasets~($3$~Hz for 360Loc and $24$ Hz for our custom dataset).

We evaluate the localization results using the Absolute Trajectory Error~(ATE) 
defined on $\mathrm{SE}(3)$.
At query frame $i$, let $\mathbf{T}_i^{\mathrm{est}}$ and
$\mathbf{T}_i^{\mathrm{gt}}$ be the estimated and ground-truth camera poses,
respectively. 
For a sequence of $N$ query frames, we define the ATE residual at frame $i$ as
\begin{equation}
  \mathrm{ATE}_i
  = \left\|
    \left[\operatorname{Log}\!\left(
      (\mathbf{T}_i^{\mathrm{gt}})^{-1}
      \mathbf{T}_i^{\mathrm{est}}\right)\right]^{\vee}
    \right\|_2.
  \label{eq_ate_se3}
\end{equation}
This criterion combines the translational and rotational pose error. 
We report the \textit{mean} and \textit{median} of $\{\mathrm{ATE}_i\}_{i=1}^{N}$ 
for each scene sequence.

\subsection{Evaluation: Localization Performance}
We evaluated the localization performance on both datasets, and the results are shown 
in Table~\ref{tab_quantitative_performance}. 
Since the baseline methods (HLoc and ACE0) perform 
localization using independent images, 
we disabled our query-query process and excluded IMU measurements for a fair comparison. 
This degrades our approach to a single-frame localization setup 
(denoted as \textbf{Ours-NoSeq} in Table~\ref{tab_quantitative_performance}). 
Under this setting, our method outperforms all baseline methods.
While HLoc is operational, its metric-level localization accuracy is compromised 
due to the visual pre-processing relying on single limited-view images (Opt) 
or the ambiguities in multi-view fusion (Fus).
ACE0 exhibits severe overfitting to the map scene representation.
It breaks down entirely under large appearance gaps between mapping and localization 
(day vs. night) or in environments with low discriminability (the \textit{Avenue} scene).
Furthermore, by incorporating inter-frame information, our full pipeline (\textbf{Ours-Full}) 
significantly reduces the mean ATE, yielding stabler localization and higher accuracy 
compared to the surround-view-based Stella SLAM, even with a higher input image frequency.

\begin{figure}
  \centering
  \subfloat[In the initialization stage, the scene suffers from heavy perceptual aliasing, 
  but our system effectively detects and filters out the erroneous constraints.]{
    \centering
    \includegraphics[width=0.97\columnwidth]{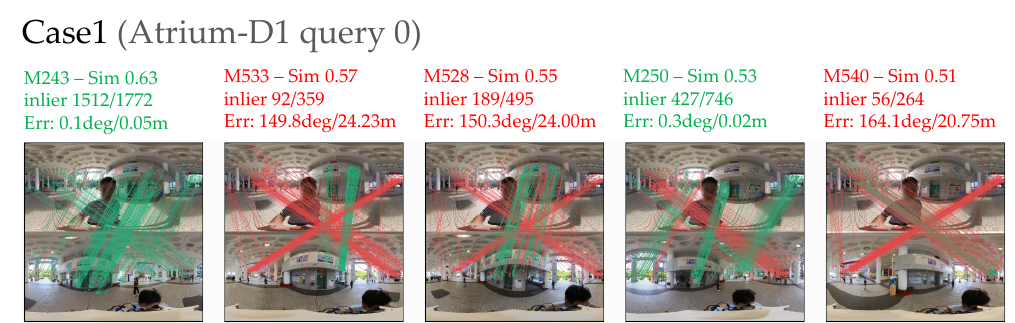}
    \label{fig_qualitative_subfig1}} \\
  \vspace{-0.2cm}
  \subfloat[Under concurrent spatial and appearance variations, our system provides 
  only one reasonable observation. It falls back to single-view matching,
  leading to a significant drop in accuracy.]{
    \centering
    \includegraphics[width=0.97\columnwidth]{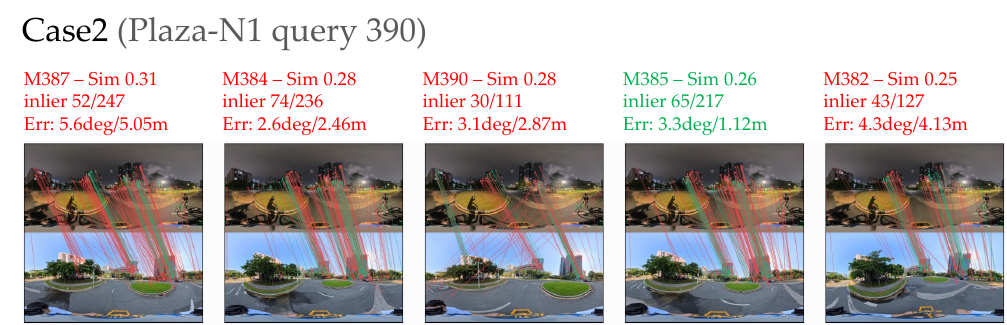}
    \label{fig_qualitative_subfig2}}
  \caption{Qualitative cases of our multiple reference localization.
  The text overlays indicate the retrieved map node ID, descriptor similarity, 
  number of RANSAC inliers, and the single-frame PnP pose error, 
  with green/red denoting accepted/rejected candidates by our verification module.
  }
  \label{fig_qualitative}
\end{figure}

Qualitative analysis~(Fig.~\ref{fig_qualitative_subfig1}) reveals that the campus scenario 
exhibits severe perceptual aliasing, posing significant challenges for global localization. 
To address this, our system employs top $N_{\mathrm{ret}}$ map image candidates 
for matching and performs multiple consistency verifications to filter out erroneous 
observation pairs, thereby significantly improving robustness.

\subsection{Evaluation: Ablation Study}
We conducted an ablation study to evaluate our system. 
By disabling the temporal processing and consistency verification modules, 
we observed that while the median ATE remains largely unaffected in most sequences, 
the mean ATE increases significantly. 
This demonstrates that incorporating temporal odometry and temporal cross-validation 
effectively enhances the stability and reliability of our system.

We also evaluated the effect of the number of map candidates ($N_{\mathrm{ret}}$) 
on accuracy, latency, and the number of candidate usage, detailed in Table~\ref{tab_ablation_top_n}.
Meanwhile, using only the best VPR result ($N_{\mathrm{ret}}=1$), as in traditional 
visual localization, allows almost all frames to pass verification but yields poor 
localization precision due to insufficient data checks from a single observation. 
Increasing the observations (e.g., $N_{\mathrm{ret}}=3$) rapidly improves accuracy, 
proving the effectiveness of our multi-observation constrained factor graph.
We find $N_{\mathrm{ret}}=5$ to be optimal by balancing the accuracy and map usage. 
Additional observations neither increase the verified candidates nor improve accuracy. 

Finally, we evaluate the computational efficiency of our system.
Accelerated via C++ and ONNX, our system operates stably at $1.30$~Hz, satisfying 
a real-time requirement of $1$~Hz for online deployment as defined 
in our system setup.

\subsection{Evaluation: Map Property}
\begin{figure}
  \centering
  \includegraphics[width=0.95\linewidth]{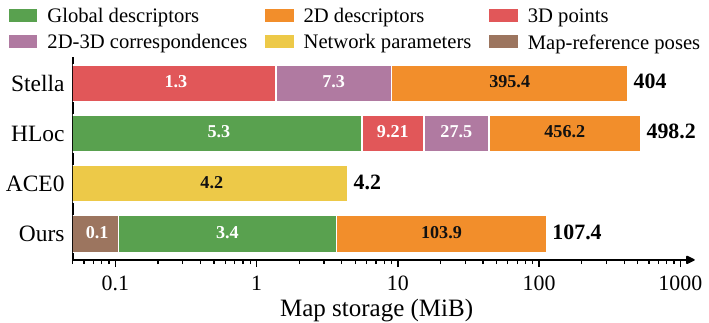}
  \caption{Visualization of map storage at 360Loc Atrium scene.
  Note that the map components are sorted by size and the x-axis is in logarithmic scale.}
  \label{fig_map_storage}
\end{figure}

We evaluated the memory footprints of the maps generated by each method, 
as visualized in Fig.~\ref{fig_map_storage}. 
Aside from ACE0's implicit neural network map ($4.2$~MB for neural parameters), 
our method consumes only $21.6\%$ and $26.6\%$ of the storage required by HLoc 
and Stella VSLAM, respectively.
Additionally, our method naturally excels at incremental mapping and map updating. 
By avoiding the triangulation of map frames and the storage of 3D point clouds, 
we structure the map as a \textit{key-value} database: 
global descriptors act as \textit{keys} for VPR queries, while 2D features and 
absolute poses serve as \textit{values}. 
This allows for straightforward map updates and expansions without the burden of 
decoupling co-visibility graphs.

\begin{table}[t]
  \caption{Comparison of top-$N_{\mathrm{ret}}$ Retrieval on computation cost,
    localization accuracy, and map-reference usage at 360Loc Atrium D1.}
  \label{tab_ablation_top_n}
  \centering
  \setlength{\tabcolsep}{2pt}
  \renewcommand{\arraystretch}{1.1}
  \newlength{\ablationcmidrulegap}
  \setlength{\ablationcmidrulegap}{\aboverulesep}
  \addtolength{\ablationcmidrulegap}{\cmidrulewidth}
  \addtolength{\ablationcmidrulegap}{\belowrulesep}
  \resizebox{\columnwidth}{!}{%
  \begin{tabular}{@{}cccccccc@{}}
    \toprule
    \multirow[c]{2}{*}[-.5\ablationcmidrulegap]{$N_{\mathrm{ret}}$}
      & \multicolumn{4}{c}{Time Consumption (ms)}
      & \multicolumn{2}{c}{ATE}
      & \multirow[c]{2}{*}[-.5\ablationcmidrulegap]{\shortstack[c]{Avg. Map-ref \\ Usage per Query}} \\
    \cmidrule(lr){2-5}\cmidrule(lr){6-7}
       & Frontend & Sph. PnP & BA    & Total   & RMSE  & Median & \\ \midrule
    1  & 300.3    & 109.8    & 20.0  & 483.7   & 1.434 & 1.258  & 0.98 \\
    3  & 410.3    & 121.3    & 31.3  & 618.1   & 0.096 & 0.057  & 2.21 \\
    5  & 521.7    & 140.4    & 49.8  & 765.9   & 0.079 & 0.054  & 3.23 \\
    8  & 685.2    & 166.9    & 73.6  & 979.6   & 0.076 & 0.055  & 4.42 \\
    10 & 798.2    & 201.6    & 76.7  & 1130.9  & 0.076 & 0.058  & 4.98 \\
    \bottomrule
  \end{tabular}%
  }
\end{table}

\section{Discussion}
We also identified several key findings during our work. 
First, although foundation models like DINOv2~\cite{oquab2023dinov2} endow current 
VPR methods with high precision, their recall remains relatively low, which inherently 
constrains the success rate of downstream metric-level localization. 
While OmniVLoc leverages this high-precision low-recall nature, 
integrating a more robust VPR paradigm, such as sequential VPR, could further boost 
global localization performance in our framework.

Second, local feature matching failures remain the primary cause of localization 
degradation in extreme cross-illumination scenarios,
particularly when coupled with severe viewpoint shifts (as illustrated 
in Fig.~\ref{fig_qualitative_subfig2}). 
Although recent work~\cite{jiao2025icra} has introduced feed-forward neural networks 
(e.g., Mast3r~\cite{leroy2024eccv}) for wide-baseline and cross-illumination matching, 
these models are typically trained on perspective camera models and struggle 
to generalize to equirectangular projections.
Therefore, designing feed-forward networks specifically tailored for omnidirectional 
cameras represents a promising direction to further bolster system robustness in 
challenging environments.

\section{Conclusion}
In this work, we proposed OmniVLoc, a topo-metric omnidirectional localization system. 
We constructed a scale-optimized visual localization paradigm driven 
by multi-view map constraints and monocular depth estimation. 
By performing pose graph estimation and precise BA optimization, our approach yields 
accurate metric-level localization. 
A hierarchical verification mechanism rejected unreliable intermediate 
outputs, ensuring high system stability. 
Experimental results confirmed that our system maintains robust localization 
despite significant illumination and scene variations.
Future work will focus on the incremental construction and lifelong maintenance of 
topo-metric omnidirectional maps, aiming for seamless compatibility with 
large-scale street-view applications.

\section*{Acknowledgment}
The authors used Gemini-3.1-Pro for text polishing and grammatical correction. 
After using this tool, the authors thoroughly reviewed and edited the 
manuscript, and take full responsibility for the final content of this 
publication.

\bibliographystyle{IEEEtran}
\bibliography{ref}

@article{yu2026tase,
  author   = {Yu, Jingwen and Yang, Jiayi and Jiao, Jianhao and Hu, Anjun and Liu, Zhonghang and Wang, Jiankun and Tan, Ping and Zhang, Hong},
  journal  = {IEEE Transactions on Automation Science and Engineering},
  title    = {ROVER: Robust Loop Closure Verification With Trajectory Prior in Repetitive Environments},
  year     = {2026},
  volume   = {23},
  number   = {},
  pages    = {8476-8489},
  doi      = {10.1109/TASE.2026.3684100}
}

@article{yin2025tro,
  author   = {Yin, Peng and Zhao, Shiqi and Wang, Jing and Ge, Ruohai and Ji, Jianmin and Hu, Yeping and Liu, Huaping and Han, Jianda},
  journal  = {IEEE Transactions on Robotics},
  title    = {iLoc: An Adaptive, Efficient, and Robust Visual Localization System},
  year     = {2025},
  volume   = {41},
  number   = {},
  pages    = {2709-2726},
  doi      = {10.1109/TRO.2025.3530273}
}

@inproceedings{camposeco2019cvpr,
  title     = {Hybrid scene compression for visual localization},
  author    = {Camposeco, Federico and Cohen, Andrea and Pollefeys, Marc and Sattler, Torsten},
  booktitle = {Proceedings of the IEEE/CVF Conference on Computer Vision and Pattern Recognition},
  pages     = {7653--7662},
  year      = {2019}
}

@article{toft2020tpami,
  title     = {Long-term visual localization revisited},
  author    = {Toft, Carl and Maddern, Will and Torii, Akihiko and Hammarstrand, Lars and Stenborg, Erik and Safari, Daniel and Okutomi, Masatoshi and Pollefeys, Marc and Sivic, Josef and Pajdla, Tomas and others},
  journal   = {IEEE Transactions on Pattern Analysis and Machine Intelligence},
  volume    = {44},
  number    = {4},
  pages     = {2074--2088},
  year      = {2020},
  publisher = {IEEE}
}

@article{yin2025tro2,
  title     = {General place recognition survey: Towards real-world autonomy},
  author    = {Yin, Peng and Jiao, Jianhao and Zhao, Shiqi and Xu, Lingyun and Huang, Guoquan and Choset, Howie and Scherer, Sebastian and Han, Jianda},
  journal   = {IEEE Transactions on Robotics},
  year      = {2025},
  publisher = {IEEE}
}

@article{campos2021tro,
  author   = {Campos, Carlos and Elvira, Richard and Rodríguez, Juan J. Gómez and M. Montiel, José M. and D. Tardós, Juan},
  journal  = {IEEE Transactions on Robotics},
  title    = {ORB-SLAM3: An Accurate Open-Source Library for Visual, Visual–Inertial, and Multimap SLAM},
  year     = {2021},
  volume   = {37},
  number   = {6},
  pages    = {1874-1890},
  doi      = {10.1109/TRO.2021.3075644}
}

@article{saputra2018csur,
  title     = {Visual SLAM and structure from motion in dynamic environments: A survey},
  author    = {Saputra, Muhamad Risqi U and Markham, Andrew and Trigoni, Niki},
  journal   = {ACM Computing Surveys (CSUR)},
  volume    = {51},
  number    = {2},
  pages     = {1--36},
  year      = {2018},
  publisher = {ACM New York, NY, USA}
}

@inproceedings{wang2021iccv,
  title     = {Continual learning for image-based camera localization},
  author    = {Wang, Shuzhe and Laskar, Zakaria and Melekhov, Iaroslav and Li, Xiaotian and Kannala, Juho},
  booktitle = {Proceedings of the IEEE/CVF International Conference on Computer Vision},
  pages     = {3252--3262},
  year      = {2021}
}

@inproceedings{maggio2023icra,
  author    = {Maggio, Dominic and Abate, Marcus and Shi, Jingnan and Mario, Courtney and Carlone, Luca},
  booktitle = {2023 IEEE International Conference on Robotics and Automation (ICRA)},
  title     = {Loc-NeRF: Monte Carlo Localization using Neural Radiance Fields},
  year      = {2023},
  volume    = {},
  number    = {},
  pages     = {4018-4025},
  doi       = {10.1109/ICRA48891.2023.10160782}
}

@article{guo2026360ral,
  title     = {360DVO: Deep Visual Odometry for Monocular 360-Degree Camera},
  author    = {Guo, Xiaopeng and Xu, Yinzhe and Huang, Huajian and Yeung, Sai-Kit},
  journal   = {IEEE Robotics and Automation Letters},
  year      = {2026},
  publisher = {IEEE}
}

@inproceedings{huang2024cvpr,
  title        = {360loc: A dataset and benchmark for omnidirectional visual localization with cross-device queries},
  author       = {Huang, Huajian and Liu, Changkun and Zhu, Yipeng and Cheng, Hui and Braud, Tristan and Yeung, Sai-Kit},
  booktitle    = {2024 IEEE/CVF Conference on Computer Vision and Pattern Recognition (CVPR)},
  pages        = {22314--22324},
  year         = {2024},
  organization = {IEEE}
}

@inproceedings{jiao2025icra,
  author    = {Jiao, Jianhao and He, Jinhao and Liu, Changkun and Aegidius, Sebastian and Hu, Xiangcheng and Braud, Tristan and Kanoulas, Dimitrios},
  booktitle = {2025 IEEE International Conference on Robotics and Automation (ICRA)},
  title     = {LiteVLoc: Map-Lite Visual Localization for Image Goal Navigation},
  year      = {2025},
  volume    = {},
  number    = {},
  pages     = {5244-5251},
  doi       = {10.1109/ICRA55743.2025.11128511}
}

@inproceedings{sarlin2019cvpr,
  title     = {From coarse to fine: Robust hierarchical localization at large scale},
  author    = {Sarlin, Paul-Edouard and Cadena, Cesar and Siegwart, Roland and Dymczyk, Marcin},
  booktitle = {Proceedings of the IEEE/CVF conference on computer vision and pattern recognition},
  pages     = {12716--12725},
  year      = {2019}
}

@inproceedings{sarlin2021cvpr,
  title     = {Back to the feature: Learning robust camera localization from pixels to pose},
  author    = {Sarlin, Paul-Edouard and Unagar, Ajaykumar and Larsson, Mans and Germain, Hugo and Toft, Carl and Larsson, Viktor and Pollefeys, Marc and Lepetit, Vincent and Hammarstrand, Lars and Kahl, Fredrik and others},
  booktitle = {Proceedings of the IEEE/CVF conference on computer vision and pattern recognition},
  pages     = {3247--3257},
  year      = {2021}
}

@inproceedings{he2024icra,
  title        = {Accurate prior-centric monocular positioning with offline lidar fusion},
  author       = {He, Jinhao and Huang, Huaiyang and Zhang, Shuyang and Jiao, Jianhao and Liu, Chengju and Liu, Ming},
  booktitle    = {2024 IEEE International Conference on Robotics and Automation (ICRA)},
  pages        = {11934--11940},
  year         = {2024},
  organization = {IEEE}
}

@inproceedings{kendall2015iccv,
  title     = {Posenet: A convolutional network for real-time 6-dof camera relocalization},
  author    = {Kendall, Alex and Grimes, Matthew and Cipolla, Roberto},
  booktitle = {Proceedings of the IEEE international conference on computer vision},
  pages     = {2938--2946},
  year      = {2015}
}

@inproceedings{brachmann2024eccv,
  title        = {Scene coordinate reconstruction: Posing of image collections via incremental learning of a relocalizer},
  author       = {Brachmann, Eric and Wynn, Jamie and Chen, Shuai and Cavallari, Tommaso and Monszpart, Aron and Turmukhambetov, Daniyar and Prisacariu, Victor Adrian},
  booktitle    = {European Conference on Computer Vision},
  pages        = {421--440},
  year         = {2024},
  organization = {Springer}
}

@inproceedings{dong2025cvpr,
  title     = {Reloc3r: Large-scale training of relative camera pose regression for generalizable, fast, and accurate visual localization},
  author    = {Dong, Siyan and Wang, Shuzhe and Liu, Shaohui and Cai, Lulu and Fan, Qingnan and Kannala, Juho and Yang, Yanchao},
  booktitle = {Proceedings of the Computer Vision and Pattern Recognition Conference},
  pages     = {16739--16752},
  year      = {2025}
}

@inproceedings{liu2025iclr,
  title     = {GS-CPR: Efficient camera pose refinement via 3d gaussian splatting},
  author    = {Liu, Changkun and Chen, Shuai and Bhalgat, Yash and Hu, Siyan and Cheng, Ming and Wang, Zirui and Prisacariu, Victor and Braud, Tristan},
  booktitle = {International Conference on Learning Representations},
  volume    = {2025},
  pages     = {76761--76780},
  year      = {2025}
}

@inproceedings{izquierdo2024cvpr,
  title     = {Optimal transport aggregation for visual place recognition},
  author    = {Izquierdo, Sergio and Civera, Javier},
  booktitle = {Proceedings of the ieee/cvf conference on computer vision and pattern recognition},
  pages     = {17658--17668},
  year      = {2024}
}

@inproceedings{berton2025cvpr,
  title     = {Megaloc: One retrieval to place them all},
  author    = {Berton, Gabriele and Masone, Carlo},
  booktitle = {Proceedings of the Computer Vision and Pattern Recognition Conference},
  pages     = {2861--2867},
  year      = {2025}
}

@article{jiao2026arxiv,
  title   = {OpenNavMap: Structure-Free Topometric Mapping via Large-Scale Collaborative Localization},
  author  = {Jiao, Jianhao and Liu, Changkun and Yu, Jingwen and Liu, Boyi and Zhang, Qianyi and Wang, Yue and Kanoulas, Dimitrios},
  journal = {arXiv preprint arXiv:2601.12291},
  year    = {2026}
}

@inproceedings{wang2024iclr,
  title     = {Freereg: Image-to-point cloud registration leveraging pretrained diffusion models and monocular depth estimators},
  author    = {Wang, Haiping and Liu, Yuan and Wang, Bing and Sun, Yujing and Dong, Zhen and Wang, Wenping and Yang, Bisheng},
  booktitle = {International Conference on Learning Representations},
  volume    = {2024},
  pages     = {28484--28507},
  year      = {2024}
}

@inproceedings{huang2022360vo,
  title        = {360vo: Visual odometry using a single 360 camera},
  author       = {Huang, Huajian and Yeung, Sai-Kit},
  booktitle    = {2022 international conference on robotics and automation (icra)},
  pages        = {5594--5600},
  year         = {2022},
  organization = {IEEE}
}

@article{oquab2023dinov2,
  title   = {Dinov2: Learning robust visual features without supervision},
  author  = {Oquab, Maxime and Darcet, Timoth{\'e}e and Moutakanni, Th{\'e}o and Vo, Huy and Szafraniec, Marc and Khalidov, Vasil and Fernandez, Pierre and Haziza, Daniel and Massa, Francisco and El-Nouby, Alaaeldin and others},
  journal = {arXiv preprint arXiv:2304.07193},
  year    = {2023}
}

@inproceedings{lin2026cvpr,
  title     = {Depth any panoramas: A foundation model for panoramic depth estimation},
  author    = {Lin, Xin and Song, Meixi and Zhang, Dizhe and Lu, Wenxuan and Li, Haodong and Du, Bo and Yang, Ming-Hsuan and Nguyen, Truong and Qi, Lu},
  booktitle = {Proceedings of the IEEE/CVF Conference on Computer Vision and Pattern Recognition},
  pages     = {26835--26844},
  year      = {2026}
}

@inproceedings{leroy2024eccv,
  title        = {Grounding image matching in 3d with mast3r},
  author       = {Leroy, Vincent and Cabon, Yohann and Revaud, J{\'e}r{\^o}me},
  booktitle    = {European conference on computer vision},
  pages        = {71--91},
  year         = {2024},
  organization = {Springer}
}

@inproceedings{detone2018cvpr,
  title     = {Superpoint: Self-supervised interest point detection and description},
  author    = {DeTone, Daniel and Malisiewicz, Tomasz and Rabinovich, Andrew},
  booktitle = {Proceedings of the IEEE conference on computer vision and pattern recognition workshops},
  pages     = {224--236},
  year      = {2018}
}

@inproceedings{lindenberger2023iccv,
  title     = {Lightglue: Local feature matching at light speed},
  author    = {Lindenberger, Philipp and Sarlin, Paul-Edouard and Pollefeys, Marc},
  booktitle = {Proceedings of the IEEE/CVF international conference on computer vision},
  pages     = {17627--17638},
  year      = {2023}
}

@inproceedings{sumikura2019acmmm,
  title     = {OpenVSLAM: A versatile visual SLAM framework},
  author    = {Sumikura, Shinya and Shibuya, Mikiya and Sakurada, Ken},
  booktitle = {Proceedings of the 27th ACM international conference on multimedia},
  pages     = {2292--2295},
  year      = {2019}
}

@article{gao2025tase,
  title     = {FeRF-BEVIO: A Feature Radiance Field-Based Bidirectionally Enhanced Visual-Inertial Odometry},
  author    = {Gao, Yuanxi and Yuan, Jing and Yu, Shizhuo and Zhang, Xuebo},
  journal   = {IEEE Transactions on Automation Science and Engineering},
  year      = {2025},
  publisher = {IEEE}
}

@article{zhang2026sj,
  author   = {Zhang, Xuetao and Sun, Gang and Lin, Shuyue and Liu, Yisha and Zhang, Xuebo and Zhuang, Yan},
  journal  = {IEEE Sensors Journal},
  title    = {TDHVIO: A Texture Richness Guided Double-Hybrid Visual–Inertial Odometry},
  year     = {2026},
  volume   = {26},
  number   = {10},
  pages    = {14576-14585},
  doi      = {10.1109/JSEN.2025.3596370}
}

@article{vidal2018ral,
  title     = {Ultimate SLAM? Combining events, images, and IMU for robust visual SLAM in HDR and high-speed scenarios},
  author    = {Vidal, Antoni Rosinol and Rebecq, Henri and Horstschaefer, Timo and Scaramuzza, Davide},
  journal   = {IEEE Robotics and Automation Letters},
  volume    = {3},
  number    = {2},
  pages     = {994--1001},
  year      = {2018},
  publisher = {IEEE}
}

\end{document}